\documentclass[journal]{IEEEtran}
\ifCLASSINFOpdf
\else
\fi
\usepackage{cite}

\usepackage{subfig}
\usepackage{amsmath,amssymb,amsfonts}
\usepackage{dsfont}
\usepackage{algorithmic}
\usepackage{algorithm}
\usepackage{graphicx}
\usepackage{textcomp}
\usepackage{xcolor}
\usepackage{lipsum}
\usepackage{enumitem}
\usepackage{tabularx,booktabs}
\newcolumntype{C}{>{\centering\arraybackslash}X} 
\usepackage{subfig}
\def\BibTeX{{\rm B\kern-.05em{\sc i\kern-.025em b}\kern-.08em
    T\kern-.1667em\lower.7ex\hbox{E}\kern-.125emX}}

\begin{document}
%
\title{Benchmarking Online Object Trackers for Underwater Robot Position Locking Applications}
%
%
%

\author{Ali~Safa,~\IEEEmembership{Member,~IEEE}, Waqas Aman,~\IEEEmembership{Member,~IEEE}, Ali Al-Zawqari,~\IEEEmembership{Member,~IEEE}, 
Saif Al-Kuwari,~\IEEEmembership{Senior Member,~IEEE}  
        
\thanks{Ali Safa, Waqas Aman and Saif Al-Kuwari are with the College of Science and Engineering, Hamad Bin Khalifa University, Doha, Qatar (e-mail: asafa@hbku.edu.qa; waman@hbku.edu.qa; smalkuwari@hbku.edu.qa).}
\thanks{Ali Al-Zawqari is with the ELEC Department, Vrije Universiteit Brussels, 1050 Brussels, Belgium (e-mail: aalzawqa@vub.be).}
\thanks{Ali Safa supervised the project as principal investigator, designed the experiments and contributed to the technical developments. Waqas Aman designed and contributed to the experiments. Ali Al-Zawqari contributed to the technical developments. Saif Al-Kuwari provided advices during the redaction of the manuscript. All authors contributed to the writing of the manuscript.}
}

\maketitle

\begin{abstract}
Autonomously controlling the position of Remotely Operated underwater Vehicles (ROVs) is of crucial importance for a wide range of underwater engineering applications, such as in the inspection and maintenance of underwater industrial structures. Consequently, studying vision-based underwater robot navigation and control has recently gained increasing attention to counter the numerous challenges faced in underwater conditions, such as lighting variability, turbidity, camera image distortions (due to bubbles), and ROV positional disturbances (due to underwater currents). In this paper, we propose (to the best of our knowledge) a first rigorous unified benchmarking of more than seven Machine Learning (ML)-based one-shot object tracking algorithms for vision-based position locking of ROV platforms. We propose a position-locking system that processes images of an object of interest in front of which the ROV must be kept stable. Then, our proposed system uses the output result of different object tracking algorithms to automatically correct the position of the ROV against external disturbances. We conducted numerous real-world experiments using a BlueROV2 platform within an indoor pool and provided clear demonstrations of the strengths and weaknesses of each tracking approach. Finally, to help alleviate the scarcity of underwater ROV data, we release our acquired data base as open-source with the hope of benefiting future research. 
\end{abstract}
 
\begin{IEEEkeywords}
Object tracking, One-shot Machine Learning-based detection, Underwater robot control, Position locking.
\end{IEEEkeywords}
\section*{Supplementary Material}
The underwater ROV dataset acquired in this work is openly available at \texttt{https://tinyurl.com/3es8u4dn}
\section{Introduction}
\label{lintro}
\IEEEPARstart{U}{nderwater} robot navigation has gained increasing attention for applications ranging from the maintenance of underwater structures and energy plants to environmental and biodiversity monitoring applications \cite{8594445, 6347898, 10161282}. Vision-based underwater robot control is known to pose several challenges compared to on-land robot navigation \cite{joshi20243dwaterqualitymapping, 10214407}, due to environmental factors such as degraded lighting conditions, lack of high-quality visual features that can be tracked, visual distortions such as bubbles and turbidity, as well as challenging water dynamics caused by currents that further disturb the position of Remotely Operated underwater Vehicles (ROVs) \cite{7989087, 6913908, 7081165, 9775489}.

Providing vision-based underwater ROV control is crucial for achieving many useful applications such as \textit{target tracking} \cite{8715217, 9032954}, where a user-specified moving object needs to be tracked by the ROV \cite{9499961}, and \textit{position control and locking} \cite{10349373, 9624231, 8207352}, where the ROV's position must be kept as stable as possible even under strong disturbances. The latter application is crucial for scenarios such as underwater inspection and constitutes our focus in this paper.
\begin{figure}[t]
\centering
    \includegraphics[scale = 0.25]{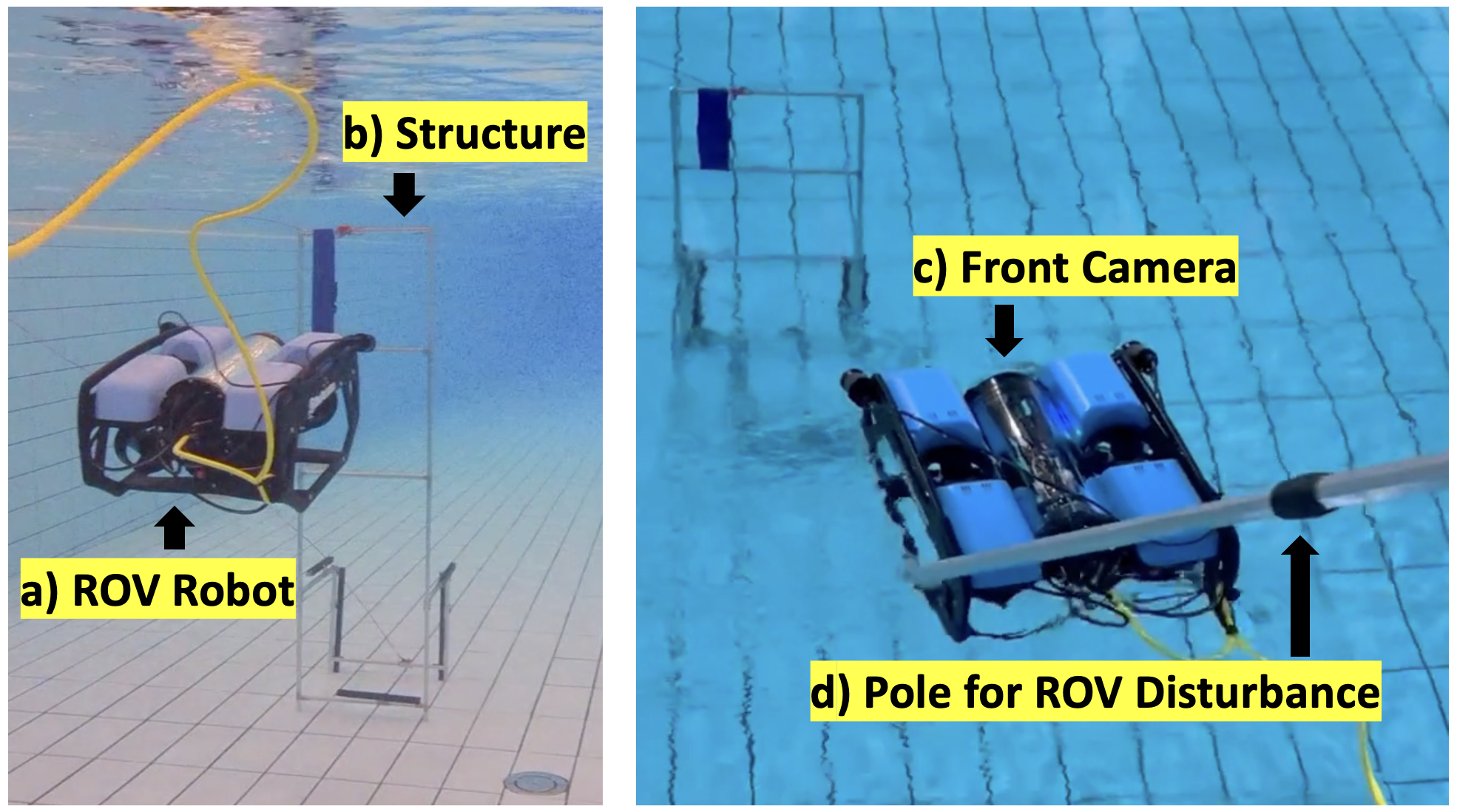}
    \caption{\textit{\textbf{Benchmarking online tracking algorithms for underwater position locking.} a) the Blue Robotics BlueROV2 robot used in our experiments; b) An underwater structure is used as an object of interest to be tracked by the ROV; c) the ROV is equipped with a front-facing camera which is used by the object tracker under test to detect and track objects as part of the position locking system; d) a pole is used to generate arbitrary disturbances on the ROV position in order to study the robustness of the object trackers during our experiments.}}
    \label{entryconcept}
\end{figure}

In this paper, we investigate the problem of underwater ROV position locking using \textit{visual-inertial fusion} \cite{9675540, 10018713} by using both camera data and the output of the ROV's Inertial Measurement Unit (IMU), which provides estimations of the ROV's yaw angle pose. We propose a system that enables an external operator to define an object of interest in front of which the ROV must remain fixed. The object of interest is then tracked by machine learning (ML)-based one-shot detection and tracking algorithms \cite{Lukezic2017}. Finally, the result of the tracking is used within a Proportional-Integrate (PI) controller \cite{10815886} to stabilize the position of the ROV under various forms of disturbances (see Fig. \ref{entryconcept}). 
In particular, since visual object tracking constitutes a crucial aspect of the proposed position locking system, this paper provides a detailed study on the performance of \textit{seven} different popular object detection and tracking algorithms \cite{grabner2006real, 5206737, 5596017, 5539960, kalal2011tracking, danelljan2014adaptive, Lukezic2017} on ROV position locking accuracy. The goal of these experiments is to provide rigorous benchmarking of a wide range of object tracking algorithms found in the literature for the specific case of underwater ROV control, with the goal of helping other researchers to quickly identify which tracking algorithm would be better suited for their own application.

The contributions of this paper can be summarized as follows:
 \begin{enumerate}
      \item We provide a unified benchmarking of seven different ML-based one-shot object detection and tracking methods for the specific context of underwater ROV position control. To the best of our knowledge, this is the first time such comprehensive benchmarking has been provided in the literature. 
      \item We describe the design of our automated ROV position locking system which integrates the output results of the different tracker algorithms assessed in this work with a Proportional-Integral (PI) controller in order to study the impact of each tracker on the real-time position locking performance.
      \item We conduct numerous real-world experiments using a real-world BlueROV2 platform to assess the performance of each tracking algorithm in terms of positional deviation error. This provides the research community with clear and comprehensive demonstrations of which tracker algorithms are better suited when designing underwater ROV control applications. 
      \item Finally, to help alleviate the scarcity of open-source underwater ROV datasets, we openly release the dataset recorded in this work as supplementary material with the hope of benefiting future research.
  \end{enumerate}
  
This rest of this paper is organized as follows. First, related works are reviewed in Section \ref{related}. Next, Section \ref{trackers} provides a technical background of the various tracker algorithms benchmarked in this paper. Then, Section \ref{control} describes the autonomous position locking control strategy used during our experiments, which takes the result of each tracker algorithm benchmarked in this work to lock the position of the ROV under various disturbances. After this, Section \ref{expsetup} describes the experimental setup used in this work while our experimental results are reported in Section \ref{expresult}. Finally, conclusions are provided in Section \ref{concsec}.

\section{Related Work}
\label{related}
Visual object tracking in underwater environments has a wide range of applications \cite{6263248, 9904898}, including monitoring animal behavior, tracking pollutants and habitat changes, exploring shipwrecks, intrusion detection, surveillance, and various industrial uses such as pipeline inspection, seabed mapping, and offshore operations. However, underwater environments present unique challenges such as poor visibility, dynamic lighting conditions, and occlusions caused by water turbidity \cite{fabbri2018underwater}. Although numerous benchmarks are available for the evaluation of object trackers in open-air environments, research on underwater-specific benchmarks remains limited \cite{9032954,Panetta:IJOE:2022, Alawode_2022_ACCV}.

In open-air environments, several benchmarks have been developed for both single-object and multi-object tracking. Among the most notable are Visual Object Tracking (VOT) \cite{kristan2015vot}, Multi Object Tracking (MOT) \cite{leal2015motchallenge} and the National University of Singapore People and Rigid Objects (NUS-PRO) tracking \cite{Li2015NUSPRO}, which have significantly helped advance the field, driving improvements in speed, precision, and success rates. In fact, object tracking is a highly important and active research area and a variety of visual object trackers have been proposed in the past few years \cite{chen2022visual}. Among the prominent and openly available Machine Learning-based one-shot trackers are Channel and Spatial Reliability Tracker \textit{(CSRT)} \cite{Lukezic2017}, Kernelized Correlation Filters \textit{(KCF)} \cite{henriques2014high}, \textit{Boosting} method \cite{Grabner2006}, Multiple Instance Learning tracker \textit{(MIL)} \cite{5206737}, Tracking-Learning-Detection method \textit{(TLD)} \cite{kalal2011tracking}, \textit{MedianFlow} \cite{5596017}, and Minimum Output Sum of Squared Error tracker \textit{(MOSSE)} \cite{5539960}. These trackers have been widely exploited in diverse computer vision applications, mostly in on-land and open-air applications \cite{dardagan2021multiple}.

Several studies \cite{9032954, Panetta:IJOE:2022, Alawode_2022_ACCV} emphasize the need to develop datasets and evaluation frameworks specifically tailored to underwater scenarios. Various one-shot visual object trackers such as \textit{KCF}, \textit{MOSSE}, \textit{Boosting}, as well as a number of deep learning (DL)-based trackers were evaluated within these object tracking benchmarks. But crucially, none of these benchmarks proposes to close the perception-control loop and to provide an assessment of the different trackers within a ROV navigation and position control context (as opposed to our study in this paper).

In addition to not assessing the tracking quality within a real-world ROV control scenario \cite{10638434}, most of these prior work focus on the use of DL-based tracker, which requires a prior offline training phase, demanding the availability and acquisition of a training data set that faithfully captures the scenarios that the ROV will encounter during its deployment underwater. In contrast, in this paper, we focus on the use of ML-based one-shot tracker algorithms that do not require any offline training phase or any tedious training data acquisition, making these methods more \textit{generalizable} to any application context and underwater environment encountered by the ROV. 


 
However, real-world and real-time benchmark experiments involving ROV platforms remains significantly unexplored. Real-time setups inherently introduce a range of complexities, including motion-induced blur from the dynamic movement of ROVs, latency in system response affecting tracking precision, and environmental disturbances particulate and bubble interference as well as underwater currents pushing the ROV platform away from its desired position. 
This paper addresses these gaps by systematically benchmarking more than seven popular ML-based one-shot trackers within numerous real-world ROV position control and locking experiments in order to directly benchmark the effect of each tracker under test with the position control precision achieved by the ROV platform.

\section{Background}
\label{trackers}

This section provides an overview of each one-shot object tracking algorithm that we benchmark in this paper and briefly details their algorithmic principles.

\subsection{Boosting Tracker}
The \textit{Boosting} Tracker employs an online version of the AdaBoost algorithm, which allows for real-time updating of classifier features during object tracking~\cite{grabner2006real}. This method is highly adaptive to changes in the object's appearance and is especially effective in distinguishing the object from its background. The core operation of the Boosting Tracker is described by the following mathematical model:
\begin{equation}
H(x) = \text{sign}\left(\sum_{t=1}^T \alpha_t h_t(x)\right)
\end{equation}
where $x$ is the input image, $h_t(x)$ represents the so-called $t$-th weak classifier, $\alpha_t$ is the weight assigned to this classifier, and $T$ is the total number of weak classifiers. These weights are dynamically adjusted based on the classifier's performance during the tracking process.
As soon as a region of interest (RoI) to track is specified, the weak classifiers are trained using image features that best differentiate the RoI from its background. The training process adapts to changes in the RoI object appearance by selecting the most discriminative features: 
\begin{equation}
h_t^* = \arg\min_{h_t} \sum_{i=1}^N w_i \times I(y_i \neq h_t(x_i))
\end{equation}
where $I$ is an indicator function that outputs 1 if the classifier's prediction $h_t(x_i)$ does not match the actual label $y_i$, and $w_i$ are dynamically adjusted weights for each successive RoI training instance $x_i$.
\subsection{Multiple Instance Learning (MIL)}
The core concept of the \textit{MIL} Tracker revolves around the notion that instead of labeling individual samples as positive or negative, samples are grouped into \emph{bags}~\cite{5206737}. 
This algorithm addresses the inherent ambiguity in sample labeling by considering sets of image patches rather than individual examples.
The MIL tracker represents the target object using a set of image features collected in bags. Each bag $B_i$ is labeled as either positive or negative:
\begin{equation}
B_i = {x_{i1}, x_{i2}, ..., x_{iN}}
\end{equation}
where $x_{ij}$ represents the $j-th$ instance in the $i-th$ bag. A bag is labeled positive if at least one instance within it contains the target object, while negative bags contain no positive instances.
Similar to the \textit{Boosting} case, the classifier $H(x)$ is constructed as a combination of weak classifiers:
\begin{equation}
H(x) = \sum_{k=1}^K \alpha_k h_k(x)
\end{equation}
where $h_k(x)$ represents the $k-th$ weak classifier, and $\alpha_k$ is its corresponding weight. Each weak classifier operates on a single feature within a so-called Haar-like feature set and is defined as:
\begin{equation}
h_k(x) = \begin{cases}
1 & \text{if } p_k v_k(x) < p_k \theta_k \\
-1 & \text{otherwise}
\end{cases}
\end{equation}
where $v_k(x)$ is the feature value, $\theta_k$ is a threshold, and $p_k$ determines the direction of the inequality.
During tracking, the algorithm maintains two types of bags: positive bags constructed around the current RoI target location, and negative bags sampled from regions further from the target.
\subsection{MedianFlow}
The MedianFlow Tracker utilizes a so-called Forward-Backward (FB) error to enhance tracking accuracy and reliability, particularly adept at detecting tracking failures~\cite{5596017}. This tracker calculates object displacements between consecutive frames using optical flow~\cite{horn1981determining}. The fundamental operation of the MedianFlow Tracker can be expressed by tracking feature points forward in time and then backward, ensuring that the start and end points align closely:
\begin{equation}
FB(T_k^f |S) = \text{distance}(T_k^f, T_k^b) = \|\mathbf{x}_t - \mathbf{\hat{x}}_t\|_2
\end{equation}
where $T_k^f$ represents the forward trajectory of a point $\mathbf{x}_t$, and $T_k^b$ represents the backward trajectory starting from $\mathbf{x}_{t+k}$ back to $\mathbf{x}_t $, where $\mathbf{\hat{x}}_t$ is the estimated initial position after backward tracking. 
The FB error is used to evaluate the consistency of the tracking, with high discrepancies indicating potential tracking failures. The tracker operates by initially detecting feature points within a user-defined RoI bounding box and tracking these points forward in time. After the forward tracking phase, the points are tracked backward to their initial positions.
The tracker performs robust filtering by automatically removing 50\% of the worst predictions based on their error measures. 
where the bounding box $\beta_{t+1}$  for the next frame $t+1$ is estimated by applying a median filter to the remaining feature points. The tracker must also account for changes in object size. It does this by computing a bounding box scale factor:
\begin{equation}
scale = \text{Median}\left(\frac{d(\mathbf{x}_{i}^{t+1}, \mathbf{x}_{j}^{t+1})}{d(\mathbf{x}_i^t, \mathbf{x}_j^t)}\right)
\end{equation}
where $d(\mathbf{x}_i, \mathbf{x}_j)$ represents the distance between the points i and j. 
\subsection{Minimum Output Sum of Squared Error (MOSSE)}
The MOSSE Tracker uses adaptive correlation filters to achieve rapid and accurate tracking performance~\cite{5539960}. These filters optimize tracking accuracy when objects change in scale, pose, and lighting conditions.
The tracker initializes its correlation filter using the initial user-selected RoI frame. This filter adapts quickly to the target object's appearance through the following formulation:
\begin{equation}
H = \frac{\sum_{i=1}^{N} \overline{G_i} \cdot F_i}{\sum_{i=1}^{N} F_i \cdot \overline{F_i}}
\label{eqmosse}
\end{equation}
where $F_i$ is the Fourier transform of the $i$-th training image, $G_i$ represents the desired output label peak indicator \textit{in the frequency domain}, and $H$ is the filter to be learned. The bar notation in (\ref{eqmosse}) indicates complex conjugation.
The filter minimizes the output sum of squared errors between estimated and actual target positions:
\begin{equation}
\min_H \sum_{i} || H \star x_i - y_i ||_2^2
\end{equation}
where $\star$ denotes correlation, $x_i$ represents training images, and $y_i$ represents desired correlation outputs.
The tracker also includes a mechanism to detect tracking failures. It uses the Peak-to-Sidelobe Ratio (PSR) to assess the quality of tracking:
\begin{equation}
PSR = \frac{peak - mean(sidelobes)}{std(sidelobes)}
\end{equation}
The PSR score measures the clarity of the correlation peak response. A low PSR score indicates potential tracking issues. The tracker can pause and resume tracking adaptively based on this score. 

\subsection{Tracking Learning Detection (TLD)}
The TLD framework has been proposed for tracking unknown objects in video streams over long periods~\cite{kalal2011tracking}. TLD divides the tracking task into three components: tracking, learning, and detection. These components work together to handle occlusions, scale changes, and varying lighting conditions.
The tracking component estimates object motion between consecutive frames. It assumes the object makes small movements and stays visible. The detection component processes each frame independently. It detects all known appearances of the object and corrects the tracker when it loses the target.
The learning component introduces a novel P-N learning approach with so-called P-expert and N-expert models complementary working together to provide detection: 
\begin{equation}
 \text{P-N Learning} = \begin{cases}
\text{P-expert:} & \text{estimates missed detections} \\
\text{N-expert:} & \text{estimates false alarms}
\end{cases}
\end{equation}
The P-expert identifies missed detections whereas the N-expert identifies false alarms. This learning process is modeled as a discrete dynamical system, ensuring that the detector progressively improves and adapts to avoid previous errors and adapts to new object appearances. 
TLD combines the outputs from its tracker and detector components in real time to continuously update the detection model. This design allows TLD to track objects that may temporarily leave the camera's view. The framework shows significant improvements over traditional tracking methods, particularly in handling object disappearance and reappearance.
\subsection{Kernelized Correlation Filters (KCF)}
The KCF Tracker proposes an alternative tracking-by-detection method through the use of efficient Fast Fourier Transforms (FFT)~\cite{duhamel1990fast} in kernelized space~\cite{danelljan2014adaptive}. By using multi-dimensional color features and adaptive dimensionality reduction, KCF maintains accuracy amidst changes in object appearance. The tracker processes so-called sub-windows (sub-frames) from each input camera frame using circulant matrix structures for rapid correlation in the frequency domain:
\begin{equation}
H = \frac{\sum_{i=1}^{N} F_i \overline{Y_i}}{\sum_{i=1}^{N} F_i \overline{F_i} + \lambda}
\end{equation}
where $F_i$ is the FFT of the $i$-th training subwindow, $Y_i$ represents the desired output, $\overline{F_i}$ is the complex conjugate of $F_i$, and $\lambda$ is a regularization parameter. KCF incorporates color attributes and kernel methods to enhance object discrimination:
\begin{equation}
\text{Response} = \mathcal{F}^{-1} \left( H \odot \mathcal{F}(z) \right)
\end{equation}
where $\mathcal{F}$ and $\mathcal{F}^{-1}$ represent the Fourier and inverse Fourier transforms, respectively, $z$ is the current image patch, and $\odot$ denotes element-wise multiplication of the detection filter $H$. KCF employs an adaptive update scheme to adjust to appearance changes, efficiently updating the model without the need to store all previous appearances. The update process is defined as:
\begin{align}
A_p^N &= (1 - \gamma) A_{p-1}^N + \gamma Y_p U_p^x, \\
A_p^D &= (1 - \gamma) A_{p-1}^D + \gamma U_p^x (U_p^x + \lambda),
\end{align}
where $A_p^N$ and $A_p^D$ are the numerator and denominator of the updated classifier filter model, respectively, $\gamma$ is the learning rate, and $U_p^x$ is the Fourier transformed kernel output for the current frame. This update scheme enhances the model's responsiveness to new information while maintaining computational efficiency, enabling real-time tracking performance.
\subsection{Channel and Spatial Reliability Tracker (CSRT)}
\label{csrt}
CSRT~\cite{Lukezic2017} represents a popular algorithm within the class of discriminative correlation filter (DCF) tracking methods~\cite{5539960}. This algorithm introduces two key innovations: channel reliability weighting and spatial reliability mapping. At its core, CSRT employs a correlation filter $h$ that learns from user-selected RoI training samples $x$ to predict object locations. The basic correlation operation can be expressed as:
\begin{equation}
f(x) = h \star x
\end{equation}
where $\star$ denotes the correlation operation.  Channel reliability is used to adjust the response of each channel during the localization stage. 
Each channel's filter $h_c$ within the filter bank $h$ is optimized individually:
\begin{equation}
h^* = \arg\min_h \sum_{c=1}^C \left( \| h_c \star x_c - y \|^2 + \lambda \|h_c\|^2 \right)
\label{firstthi}
\end{equation}
where $x_c$ represents the feature channel $c$ of the input $x$, $y$ is the desired output, and $\lambda$ is a regularization parameter to alleviate over-fitting on the RoI that is being tracked.

CSRT also introduces the so-called spatial reliability map $m(x)$ into (\ref{firstthi}) which modulates the learning process to emphasize more on the reliable parts of the target. By introducing $m(x)$, the optimization process is re-defined as:
\begin{equation}
h^* = \arg\min_h \|m(x) \odot (h \star x - y)\|^2 + \lambda \|h\|^2
\end{equation}
where $m(x)$ is the spatial reliability map that applies weights at each spatial location, enhancing focus on regions suitable for tracking, and $\odot$ denotes element-wise multiplication. CSRT follows an iterative optimization process where the tracker updates channel weights and the spatial reliability map based on observed changes in the target's appearance. This adaptive approach is crucial for handling object deformation and partial occlusions effectively. 

Crucially, all the methods covered in the Section are one-shot learning algorithms that do not require any model pre-training as in the case of Deep Learning-based techniques (which are not the scope of this work). Instead, the methods covered above only require an initial RoI selected by the user and indicating the object to be tracked within the scene. Then, the algorithms listed above keep on tracking the specified object while updating their detection model using the successively tracked RoIs that were detected within the camera frames. This enables these methods to still generalize well to any unseen environment in an agnostic manner, and perform without the need for tedious training set acquisition and labeling (which is needed to capture each specific environment in which the algorithm must be deployed).

\section{Autonomous ROV position locking pipeline}
\label{control}
This Section describes the ROV autonomous position control strategy adopted in this work. The position control pipeline takes as input the tracking result provided by the different tracker algorithms under test, and uses the tracked bounding box coordinates alongside a Proportional-Integral (PI) controller to regulate the position of the ROV. First, the pre-processing steps applied on the tracker results are detailed in Section \ref{preprocessbbox}. Then, the PI controller taking as input the result of the pre-processing scheme is described in Section \ref{univariatePID}. 

\subsection{Pre-processing Steps}
\label{preprocessbbox}
We use the camera data $C_k$ and the yaw angle measurement $\Psi_k$ as input for the controller pipeline proposed in this work (where $k$ denotes the discrete-time index).

In order to lock the position of the ROV with regard to a specific object that needs to be maintained within the ROV's field of view, an external operator can select an object of interest on the live camera feed $C_k$ by drawing a \textit{bounding box} around the target object. We denote the coordinates of the user-specified bounding box as $(x_u,y_u,w_u,h_u)$ where $x_u,y_u$ denote the corner origin coordinate of the bounding box on the image plane, and $w_u,h_u$ respectively denote its width and hight.

Then, once the user-specified bounding box is obtained, the locking control procedure begins by first detecting the user-specified object on the next camera frame $C_{k+1}$ and then using the new bounding box $(x_k,y_k,w_k,h_k)$ detected on the previous frame $C_k$ to measure in real-time the x-axis deviation $\Delta X$, y-axis deviation $\Delta Y$ and scale deviation $\Delta S$ between the current bounding box position of the object and the original user-specified bounding box $(x_u,y_u,w_u,h_u)$:
\begin{equation}
    \Delta X = \frac{x_u+w_u}{2} - \frac{x_k+w_k}{2}
    \label{eqx}
\end{equation}
\begin{equation}
    \Delta Y = \frac{y_u+h_u}{2} - \frac{y_k+h_k}{2}
    \label{eqy}
\end{equation}
\begin{equation}
    \Delta S = \frac{w_u - w_k}{2} + \frac{h_u - h_k}{2}
    \label{eqs}
\end{equation}

The object detection and tracking takes as initial point the region of interest (RoI) specified by the user-defined bounding box and is implemented using the trackers covered in Section \ref{trackers}. During our experiments, each one of the tracker algorithms covered in Section \ref{trackers} will be tried out to assess the impact of each tracking method on the position locking performance of the ROV.

In Equations (\ref{eqx}, \ref{eqy}, \ref{eqs}), $\Delta X$ and $\Delta Y$ respectively provide an indication of how much the ROV has deviated along its x- and y-axis (\textit{sway} and \textit{heave} axis), while the scale deviation $\Delta S$ provides information on how much the ROV has deviated along its \textit{surge} or z-axis (i.e., getting closer or farther from the target object).

In addition, we also compute the yaw angle deviation between the current yaw angle $\Psi_k$ and the yaw angle $\Psi_u$ that the ROV exhibited at the time of the user-defined object selection (measured using the IMU semsor on board of the ROV):
\begin{equation}
    \Delta \Psi = \Psi_k - \Psi_u
    \label{eqsp}
\end{equation}

\subsection{Proportional-Integral (PI) controller}
\label{univariatePID}
We set up a Proportional-Integral (PI) controller for each of the control axis of the ROV as follows:
\begin{multline}
    u_{\{X,Y,Z,\Psi\}} = K_p^{\{X,Y,Z,\Psi\}} \times \Delta \{X,Y,Z,\Psi\}_k 
    \\
    + K_i^{\{X,Y,Z,\Psi\}} \times \sum_{k=0}^{k^*} \Delta \{X,Y,Z,\Psi\}_k 
    \label{univarPID}
\end{multline}
where $k^*$ denotes the current time index, and where $K_p, K_i$ are respectively the proportional and integral coefficients. The indices ${\{X,Y,Z,\Psi\}}$ indicate that each of the $4$ available motor thrust axis is controlled by its own independent PI controller using the associated deviation corresponding to the thrust axis. The PI parameters $K_p^{\{X,Y,Z,\Psi\}}, K_i^{\{X,Y,Z,\Psi\}}$ are tuned empirically by visually assessing the behavior of the ROV under random position disturbances imposed during the tuning process. Finally, $u$ denotes the controller output command sent to the ROV.

In the next Section, the automated object tracking and position control strategy described above will be integrated with the BlueROV2 platform as part of the setup used during our experiments. 

\section{Experimental Setup}
\label{expsetup}

\begin{figure}[htbp]
\centering
    \includegraphics[scale = 0.35]{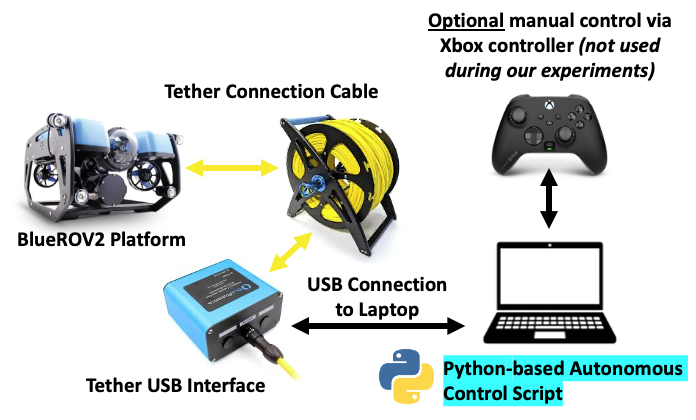}
    \caption{\textit{\textbf{Block diagram of the ROV control setup.} The ROV platform is connected via a tether to a USB interface. The laptop running the position control script receives the ROV camera feed and IMU readings via the tether USB interface and sends control commands to the ROV using this same interface. Optionally, an Xbox controller can be used for the manual control of the ROV for initiating its position in the pool (not used during our automated control experiments). }}
    \label{blockscem}
\end{figure}

We perform our experiments in an indoor pool using a BlueROV2 underwater robot platform connected to a control laptop via a tether connection. The control laptop enables both a manual control (using an Xbox controller) and an \textit{automated} control of the ROV using a \textit{Python} environment that communicates with the ROV through a \textit{MAVlink} interface. To conduct the experiments, we implement our position locking algorithm described in Section \ref{control} within a \textit{Python} script, which communicates with the ROV to acquire camera and IMU data, and sends back motor control commands for correcting the ROV's position when deviations occur. Fig. \ref{blockscem} provides a block diagram of the setup.

\begin{figure}[htbp]
\centering
    \includegraphics[scale = 0.32]{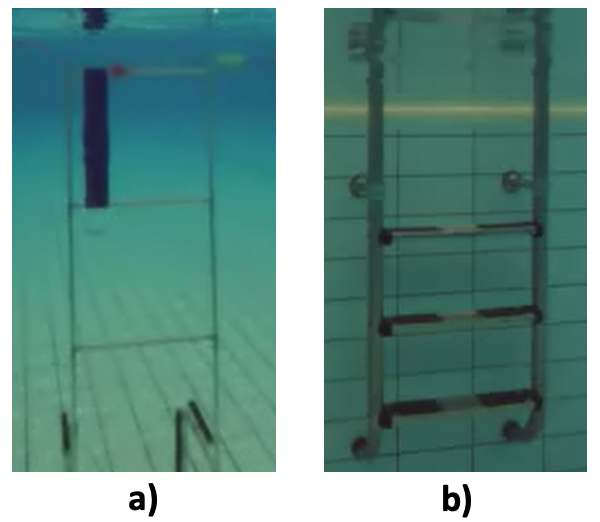}
    \caption{\textit{\textbf{Types of objects to be tracked for position locking.} a) \texttt{structure} object b) \texttt{ladder} object.}}
    \label{typesofobject}
\end{figure}

As objects to be tracked for the ROV position locking, we consider \textit{a)} a custom \texttt{structure} object built using \textit{MakerBeam} aluminum beams and \textit{b)} a \texttt{ladder} object mounted along the pool's wall (see Fig. \ref{typesofobject}). Using both of these structures during our experiments adds additional variability as the algorithm's accuracies will be assessed on two different objects with different properties in terms of location in the pool and visibility.

Using each structure, we perform three series of experiments:
\begin{enumerate}
    \item \textbf{With no disturbances:} In this first series of experiment, we are interested in studying the natural drift of each tracker algorithm caused by both visual tracking errors and the natural movement of the water in the pool.
    \item \textbf{With bubbles in front of the ROV's camera:} In this second series of experiments, we assess the robustness of each tracker to visual distortions caused by \textit{bubbles} generated using the disturbance pole (see Fig. \ref{entryconcept} d). In addition, since the disturbance pole is agitated in front of the ROV's camera, this second series of experiments also provides an assessment of the tracker algorithms accuracy under partial object \textit{occlusion} (caused by the pole that is agitated in between the object to be tracked and the ROV's camera).
    \item  \textbf{With positional disturbances:} In this last series of experiments, the disturbance pole is used to cause deviations along the yaw angle, x-axis, y-axis and z-axis position of the ROV. These disturbance deviations are meant to be large-scale so as to make the visual tracking and position locking problem more challenging ($\sim 30^\circ$ for the yaw angle and $\sim 30$ cm for the x,y,z-axis). In order to achieve systematic disturbances throughout the experiments, the disturbance sequence is kept the same across the trackers and the real-time measured deviation of the ROV due to the disturbance can be monitored using the control computer at the same time that the operator generates the ROV displacement with the pole. 
\end{enumerate}

Table \ref{dataacquisitions} summarizes the dataset acquired during the experiments conducted in this work (available as open source as supplementary material).

\begin{table}[htbp]
\centering
\begin{tabularx}{0.47\textwidth}{@{}l*{0}{c}c@{}}
\toprule
Acquisition Name  & Disturbance Type  \\ 
\midrule
no-disturb-\{s,l\}-\{\texttt{tracker\_type}\}      &  no disturbances          \\ 
bubble-disturb-\{s,l\}-\{\texttt{tracker\_type}\}      &  bubble generated using pole    \\ 
with-disturb-\{s,l\}-\{\texttt{tracker\_type}\}     &     position deviations using pole      \\ 

\bottomrule
\end{tabularx}
\caption{\textit{\textbf{ROV dataset.} In the acquisition name, the suffix -s denotes visual locking on the \texttt{structure} object while the suffix -l denotes visual locking on the \texttt{ladder} object. In addition, the suffix \texttt{tracker\_type} denotes the tracking algorithm used during the data acquisition and ROV control (see tracker types in Section \ref{trackers}).}}
\label{dataacquisitions}
\end{table}

\section{Experimental Results}
\label{expresult}

This section reports our experimental results regarding the robustness and accuracy of each tracker algorithm that is benchmarked in this paper. First, each tracker is assessed without any external disturbances in Section \ref{withoutdist} in order to provide a baseline indication of the stability of each tracker. Then, in Section \ref{withbubble}, each tracker is assessed under \textit{occlusion and bubble-type visual disturbances} generated in front of the ROV's camera. The goal of this second series of experiments is to assess to robustness of each tracker to typical bubble-type visual distortions and partial occlusions. Finally, the last series of experiments that we perform in Section \ref{withexternaldist} consists of assessing the position locking accuracy of each tracker algorithm under \textit{strong positional disturbances}, by using a pole to push the robot in arbitrary directions (see Fig. \ref{entryconcept} d).

\subsection{Without external disturbances}
\label{withoutdist}

\begin{figure}[htbp]
\centering
    \includegraphics[scale = 0.57]{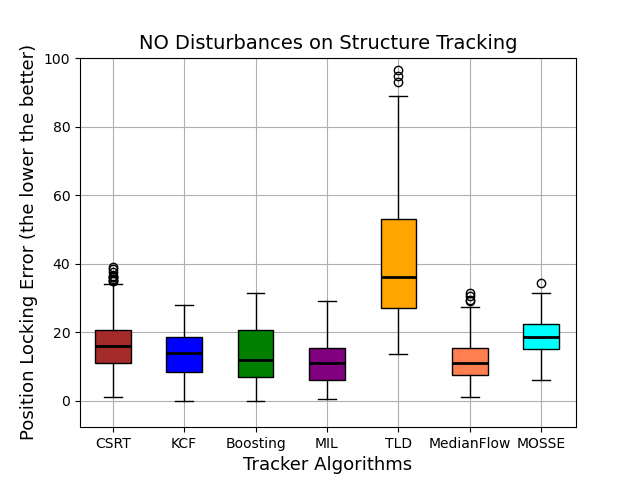}
    \caption{\textit{\textbf{Position locking error} \textbf{-} \texttt{Structure} \textbf{-} \texttt{No disturbances}. }}
    \label{res13}
\end{figure}

\begin{figure}[htbp]
\centering
    \includegraphics[scale = 0.57]{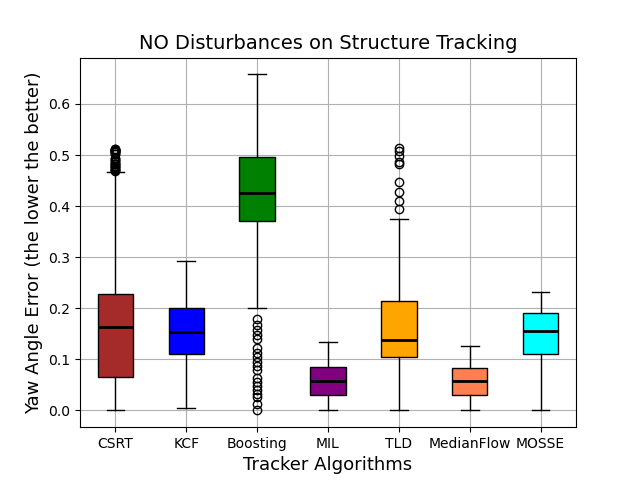}
    \caption{\textit{\textbf{Yaw angle error} \textbf{-} \texttt{Structure} \textbf{-} \texttt{No disturbances}. }}
    \label{res3}
\end{figure}

\begin{figure}[htbp]
\centering
    \includegraphics[scale = 0.57]{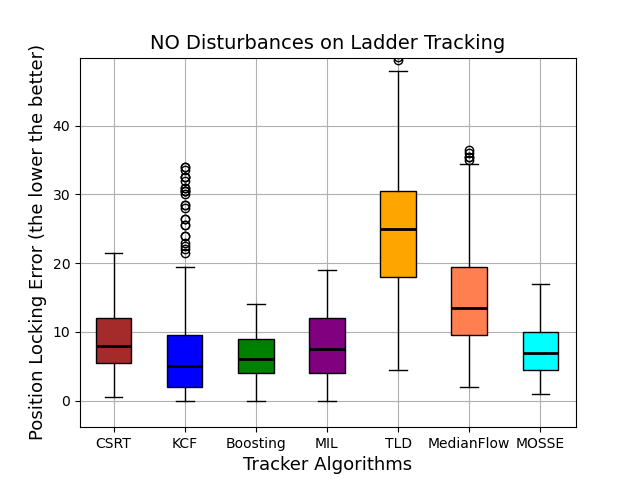}
    \caption{\textit{\textbf{Position locking error} \textbf{-} \texttt{Ladder} \textbf{-} \texttt{No disturbances}. }}
    \label{res15}
\end{figure}

\begin{figure}[htbp]
\centering
    \includegraphics[scale = 0.57]{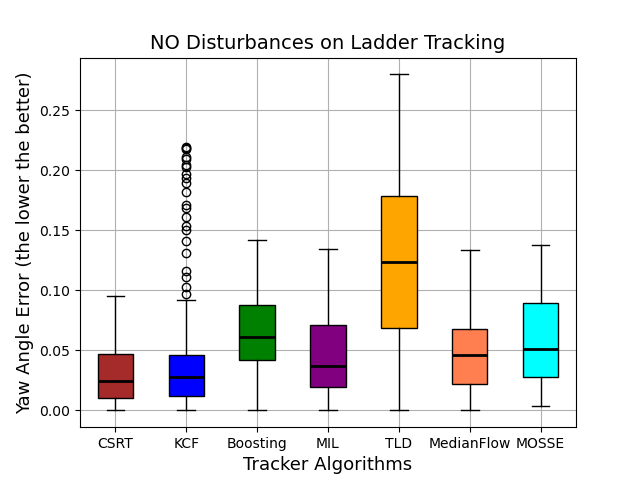}
    \caption{\textit{\textbf{Yaw angle error} \textbf{-} \texttt{Ladder} \textbf{-} \texttt{No disturbances}. }}
    \label{res5}
\end{figure}

In this first series of experiments, we are interested to study the positional locking performance of each tracker without any external disturbances other than the natural movement of the water in the swimming pool environment where the ROV operates. These experiments will enable us to gain a first assessment of the \textit{baseline} performance of each tracker algorithm. We conduct our experiments with both the \texttt{structure} and the \texttt{ladder} objects and report the obtained position locking and yaw angle deviation results in Figures \ref{res13}, \ref{res3}, \ref{res15} and \ref{res5}.

In the case of the \texttt{structure} object (Figures \ref{res13} and \ref{res3}), we remark that both the \textit{MIL} and the \textit{MedianFlow} trackers are the top performers and achieve a similar median positional and yaw angle deviation error of around $10$ pixels and $0.05$ radians respectively. On the other hand, we remark on Fig. \ref{res13} and \ref{res3} that the \textit{TLD} and \textit{Boosting} methods perform the worse, with \textit{TLD} exhibiting the highest median positional error of around $35$ pixels and with \textit{Boosting} exhibiting the highest median yaw angle error of around $0.43$ radians.  


In the case of the \texttt{ladder} object (Figures \ref{res15} and \ref{res5}), we observe that the \textit{TLD} is the worst performer, both in terms of position locking and in terms of yaw angle error. This confirms the observations previously obtained using the \texttt{structure} object in Fig. \ref{res13} where the \textit{TLD} method was also the worst performer. On the other hand, we observe in Figures \ref{res15} and \ref{res5} that the \textit{CSRT} and \textit{KCF} methods appear to be the top performers since these two methods jointly exhibit both a low positional error and a low yaw angle error (with \textit{KCF} outperforming \textit{CSRT} in terms of positional error and \textit{CSRT} outperforming \textit{KCF} in terms of yaw angle error).

Overall, we observe that the \textit{TLD} method systematically ranks among the worst-performing trackers throughout the experiments above while the \textit{KCF}, \textit{CSRT} and \textit{MIL} methods rank among the top performers (each outperforming the other methods depending on the different scenarios, either in terms of positional deviation or in terms of yaw angle deviation).

\subsection{With bubble generation in front of the camera}
\label{withbubble}

\begin{figure}[htbp]
\centering
    \includegraphics[scale = 0.35]{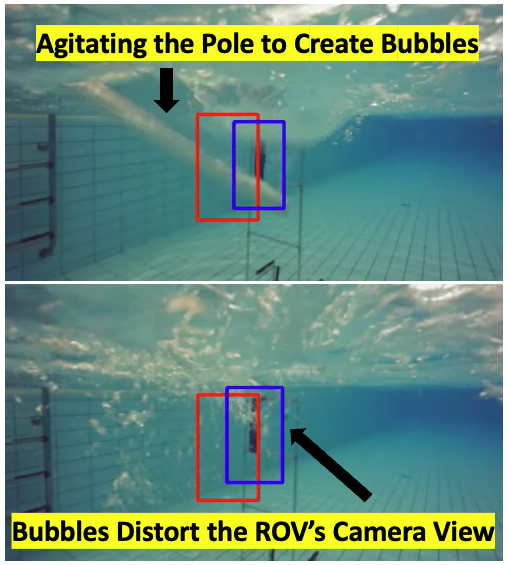}
    \caption{\textit{\textbf{Bubbles are generated} by using the pole to agitate the water in front of the ROV's camera. The blue bounding box indicates the object that is being tracked and the red bounding box shows the location and scale of the original bounding box selected by the user to which the blue bounding box must align with during the ROV control process.}}
    \label{bubblescenario}
\end{figure}

\begin{figure}[htbp]
\centering
    \includegraphics[scale = 0.57]{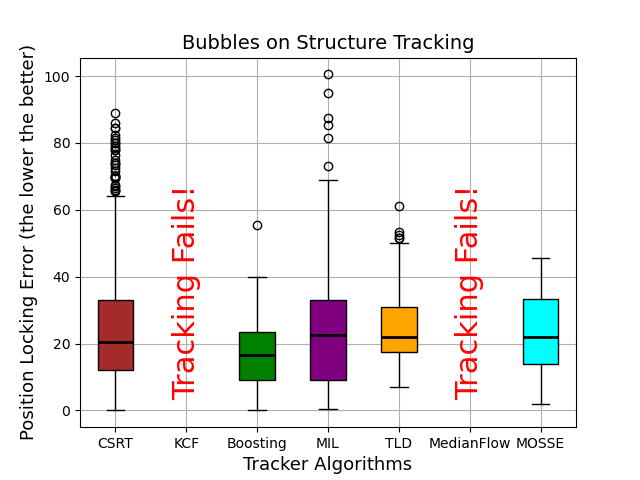}
    \caption{\textit{\textbf{Position locking error} \textbf{-} \texttt{Structure} \textbf{-} \texttt{Bubbles}. }}
    \label{res14}
\end{figure}

\begin{figure}[htbp]
\centering
    \includegraphics[scale = 0.57]{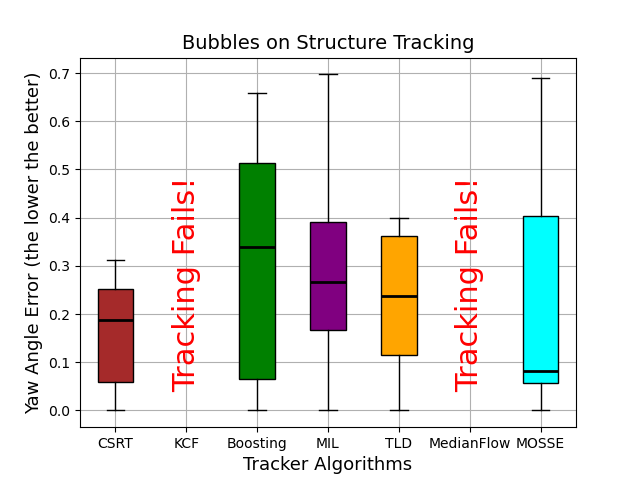}
    \caption{\textit{\textbf{Yaw angle error} \textbf{-} \texttt{Structure} \textbf{-} \texttt{Bubbles}. }}
    \label{res4}
\end{figure}

\begin{figure}[htbp]
\centering
    \includegraphics[scale = 0.57]{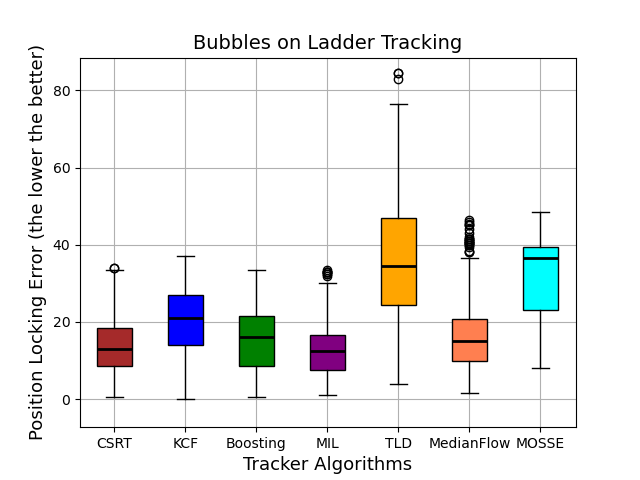}
    \caption{\textit{\textbf{Position locking error} \textbf{-} \texttt{Ladder} \textbf{-} \texttt{Bubbles}. }}
    \label{res16}
\end{figure}

\begin{figure}[htbp]
\centering
    \includegraphics[scale = 0.57]{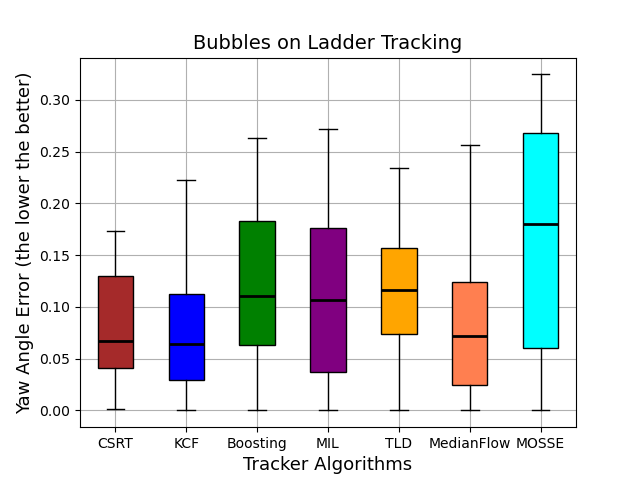}
    \caption{\textit{\textbf{Yaw angle error} \textbf{-} \texttt{Ladder} \textbf{-} \texttt{Bubbles}. }}
    \label{res6}
\end{figure}

In this second experiment, we assess the robustness of each tracking algorithm against bubble-type disturbances and partial occlusion that can occur in front of the ROV's cameras. The distortion caused by the bubbles as well as the partial occlusion caused by the agitating pole makes the target locking problem more challenging since the ROV's camera feed becomes distorted as well, leading to potential tracking errors and even the loss of track. Fig. \ref{bubblescenario} provides a view of the bubble disturbance setup considered in these experiments. We perform experiments with both the \texttt{structure} and the \texttt{ladder} objects and report the position locking and the yaw angle deviation results in Figures \ref{res14}, \ref{res4}, \ref{res16} and \ref{res6}.

In the case of the \texttt{structure} object (Figures \ref{res14} and \ref{res4}), we remark that the \textit{KCF} and \textit{MedianFlow} methods are not robust to the presence of bubbles in the camera feed and lead to a catastrophic loss of track. During our experiments, we tried both \textit{KCF} and \textit{MedianFlow} multiple times and always remarked the catastrophic track loss phenomenon due to the distortion caused by the bubbles. 

Using the \texttt{structure} object, we observe that the \textit{Boosting} method gives the best results in terms of position locking (Fig. \ref{res14}) while the \textit{CSRT} method gives the best results in terms of yaw angle error (Fig. \ref{res4}). On the other hand, we observe that the \textit{Boosting} method performs the worse in terms of yaw angle error. 


In the case of the \texttt{ladder} object (Figures \ref{res16} and \ref{res6}), we remark that all methods were able to keep their tracking, indicating that the \texttt{ladder} object is easier to distinguish and track compared to the \texttt{structure} object. In terms of position locking error, Fig. \ref{res16} shows that the \textit{MIL} and \textit{CSRT} methods provide the best results and achieve almost the same positional error.  

In terms of yaw angle error, Fig. \ref{res6} shows that the the \textit{KCF} and \textit{CSRT} methods achieve an on-par median error (see black line in the box plots) while \textit{CSRT} achieves a smaller maximum-minimum range (indicated by the T-bars in the box plots). 

Overall, the \textit{CSRT} tracker appears as the best performing method throughout the bubble disturbance experiments since it systematically ranks within the lowest reported positional and yaw angle errors, and does not suffer from track loss when using the \texttt{structure} object (as opposed to the \textit{KCF} tracker which failed when tracking the \texttt{structure} object).

\begin{table*}[!t]
\captionsetup{justification=centering}
\begin{tabularx}{\textwidth}{@{}l*{6}{C}c@{}}
\toprule
Experiments - \textbf{Position error}  & Boosting & MIL & MedianFlow & MOSSE & TLD & KCF & CSRT \\ 
\midrule
\textbf{1)} \texttt{Structure} - \texttt{No Disturbance}    & 12.1      & 10.5          & 10.6   & 19.3  & 38.4    & 17.8 & 18.1
  \\ 
\textbf{2)}  \texttt{Ladder} - \texttt{No Disturbance} & 6.7        & 7.9        & 13.4    & 7.8   & 25.1    & 5 & 8.4    \\ 
\textbf{3)}  \texttt{Structure} - \texttt{Bubbles}     & 18.6       & 21.7          & \textit{FAIL}  & 21.6  & 21.6   & \textit{FAIL} & 20.3   \\ 
\textbf{4)}  \texttt{Ladder} - \texttt{Bubbles}      & 18.1       & 16.7          & 17.9   & 38.8 & 37.9   & 20.5 & 17.3  \\ 
\textbf{5)}   \texttt{Structure} - \texttt{With disturbance}     & 20       & 30.4          & 19.1   & 19.5  & 37.2   & \textit{FAIL} & 18.2   \\
\textbf{6)}    \texttt{Ladder} - \texttt{With disturbance}    & 33.1       & \textit{FAIL}          & 17.2  & 25  & 24.9  & \textit{FAIL}& 20.6  \\
\midrule
\textbf{Average:} & 18.1 & \textit{FAIL} & \textit{FAIL} & 22 & 30.9 & \textit{FAIL} & \textbf{17.2}\\
\midrule
Experiments - \textbf{Yaw angle error} & & & & & & & \\
\midrule
\textbf{1)} \texttt{Structure} - \texttt{No Disturbance}       & 0.43        & 0.06          & 0.06   & 0.17 & 0.14 & 0.16   & 0.17  \\ 
\textbf{2)}  \texttt{Ladder} - \texttt{No Disturbance}      & 0.061        & 0.039          & 0.045  & 0.05 & 0.125  & 0.027    & 0.025  \\ 
\textbf{3)}  \texttt{Structure} - \texttt{Bubbles}      & 0.35       & 0.27         & \textit{FAIL} & 0.09  & 0.24  & \textit{FAIL}   & 0.18   \\ 
\textbf{4)}  \texttt{Ladder} - \texttt{Bubbles}         & 0.11       & 0.105         & 0.075   & 0.17 & 0.12  & 0.065    & 0.065  \\ 
\textbf{5)}   \texttt{Structure} - \texttt{With disturbance}      & 0.24       & 0.25         & 0.28  & 0.28  & 0.15  & \textit{FAIL}    & 0.15  \\ 
\textbf{6)}    \texttt{Ladder} - \texttt{With disturbance}     & 0.27       & \textit{FAIL}         & 0.25   & 0.18 & 0.37  & \textit{FAIL}    & 0.1   \\ 
\midrule
\textbf{Average:} & 0.24 & \textit{FAIL} & \textit{FAIL} & 0.16 & 0.19 & \textit{FAIL} & \textbf{0.115}\\
\bottomrule
\end{tabularx}
 \caption{\textit{\textbf{Result summary.} Both the median position error (in pixels) and the median yaw angle errors (in radians) are reported for each tracker under test. We denote by FAIL the presence of a tracking failure during the experiment.}}
\label{bigtable}
\end{table*}

\subsection{With external disturbances}
\label{withexternaldist}

In this final series of experiments, we assess the robustness of each tracking algorithm against strong positional disturbances. During our experiments, positional disturbances are applied by using the pole to cause arbitrary deviations in the yaw angle, surge (x-axis position), sway (y-axis position) and heave (z-axis position) of the ROV. The disturbance pattern is kept the same across the experiments and consists of causing \textit{i)} a yaw angle deviation in the positive direction, followed by a yaw disturbance in the negative one; \textit{ii)} a positive surge disturbance followed by a negative one; \textit{iii)} a positive sway disturbance, followed by a negative one; and \textit{iv)} a positive heave disturbance, followed by a negative one. The trained human operator manipulating the disturbance pole applies disturbances with similar amplitudes by monitoring the real-time position and yaw angle readings coming from the ROV's controller.

\begin{figure}[htbp]
\centering
    \includegraphics[scale = 0.57]{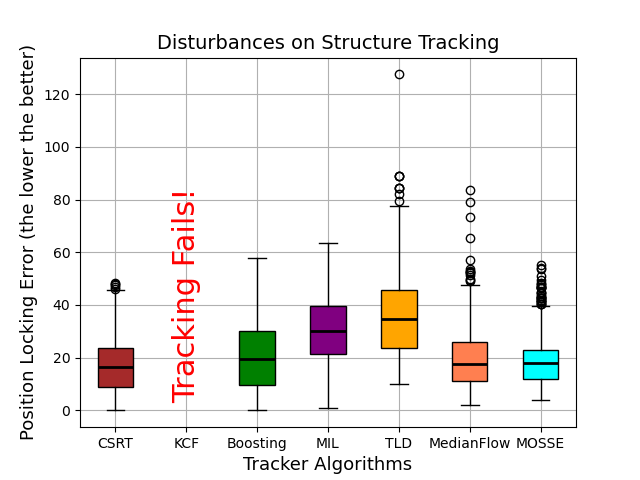}
    \caption{\textit{\textbf{Position locking error} \textbf{-} \texttt{Structure} \textbf{-} \texttt{With disturbances}. }}
    \label{res11}
\end{figure}

\begin{figure}[htbp]
\centering
    \includegraphics[scale = 0.57]{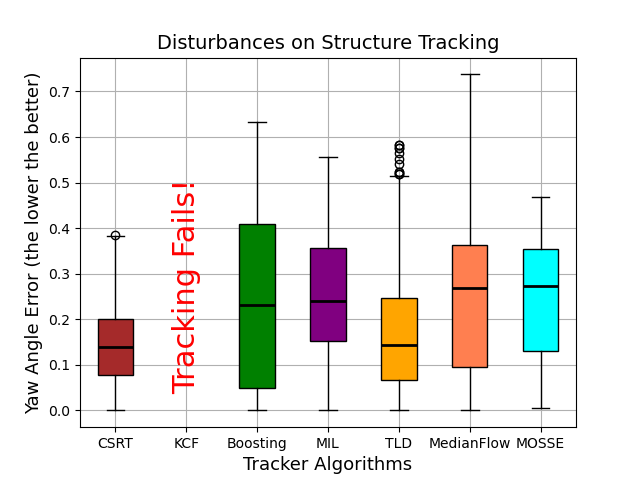}
    \caption{\textit{\textbf{Yaw angle error} \textbf{-} \texttt{Structure} \textbf{-} \texttt{With disturbances}. }}
    \label{res1}
\end{figure}

\begin{figure}[htbp]
\centering
    \includegraphics[scale = 0.57]{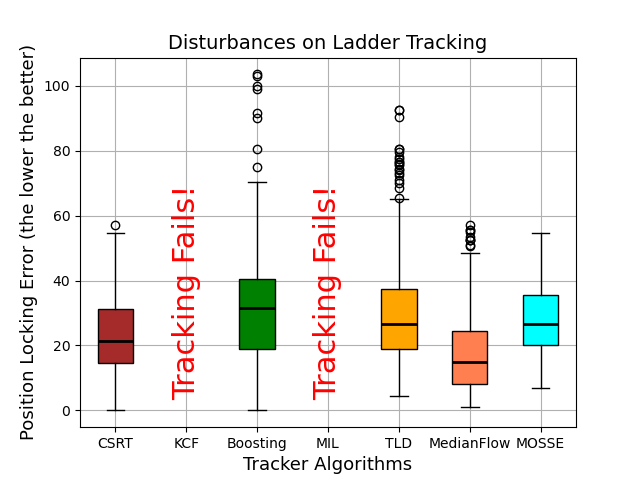}
    \caption{\textit{\textbf{Position locking error} \textbf{-} \texttt{Ladder} \textbf{-} \texttt{With disturbances}. }}
    \label{res12}
\end{figure}

\begin{figure}[htbp]
\centering
    \includegraphics[scale = 0.57]{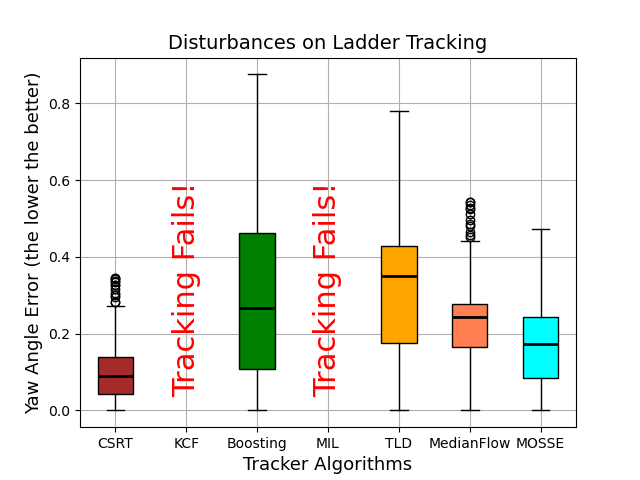}
    \caption{\textit{\textbf{Yaw angle error} \textbf{-} \texttt{Ladder} \textbf{-} \texttt{With disturbances}. }}
    \label{res2}
\end{figure}

The results obtained with each tracker on both the \texttt{structure} and \texttt{ladder} objects are provided in Figures \ref{res11}, \ref{res1}, \ref{res12} and \ref{res2}.

In the case of the \texttt{structure} object (Figures \ref{res11} and \ref{res1}), we observe that the \textit{KCF} method fails to consistently track the object, leading to a track loss (as also observed in Figures \ref{res14}, \ref{res4} of the bubble disturbance case in Section \ref{withbubble}). On the other hand, we remark in Figures \ref{res11} and \ref{res1} that the \textit{CSRT} method achieves the lowest error in terms of both positional and yaw deviation error. 

Now, with regard to the case of the \texttt{ladder} object (Figures \ref{res12} and \ref{res2}), we observe once again that the \textit{KCF} method suffers from track loss. In addition, the \textit{MIL} method also fails by suffering from track losses during the experiments. On the other hand, we remark that both the \textit{CSRT} and \textit{MedianFlow} methods appear as the best performing trackers, with \textit{MedianFlow} outperforming \textit{CSRT} in terms of positional error and CSRT outperforming \textit{MedianFlow} in terms of yaw angle error.

Overall, the \textit{CSRT} tracker appears to be the best performer throughout our positional disturbance experiments as it systematically ranks among the best methods using both the \texttt{structure} and \texttt{ladder} objects, and in terms of both positional and yaw angle deviation errors. On the other hand, the use of the \textit{KCF} and \textit{MIL} methods lead to track loss, indicating their lack of robustness against the strong positional disturbances applied to the ROV during our experiments. 

\subsection{Discussion and summary of the results}
\label{discussion}

Table \ref{bigtable} summarizes the median errors (indicated by the black lines in the box plots) obtained using each tracking algorithm for every experiment we conducted. It can be clearly seen that on average, the \textit{CSRT} tracker appears to be the top performer throughout our experiments, as it achieves the lowest average median errors of 17.2 pixels for the positional deviation and 0.115 radians for the yaw angle deviation across all experiments (see averages along all experiments in Table \ref{bigtable}). 

On the other hand, the \textit{MIL}, \textit{MedianFlow} and \textit{KCF} methods are among the lowest performers since these algorithms suffered from \textit{tracking failures} during the experiments (i.e., the object to be tracked gets lost and as a consequence, the ROV control pipeline severely drifts away).

We did not observe tracking failures using the \textit{Boosting}, \textit{MOSSE} and \textit{TLD} trackers, but all these methods reach higher average median errors compared to the \textit{CSRT} tracker, as seen in Table \ref{bigtable}. 

During our experiments, we also visually remarked the superiority of the \textit{CSRT} method by observing that the bounding boxes produced by \textit{CSRT} appear to be more stable, as opposed to the other methods which often produced bounding boxes that would suddenly change in scale in between successive frames. In addition, we observed a greater robustness with \textit{CSRT} when it came to handling partial object occlusions (e.g., during the bubble experiments). These useful properties can be linked to the key feature of \textit{channel reliability tracking} found in the \textit{CSRT} algorithm \cite{Lukezic2017}, which is crucial for handling object deformation and partial occlusion (see Section \ref{csrt}).  

Therefore, in light of the comprehensive series of real-world underwater experiments conducted in this work, it clearly appears that the \textit{CSRT} tracker is a sound candidate when designing underwater vision-based ROV control systems and should be preferably selected over the other trackers benchmarked in this paper.

\section{Conclusion}
\label{concsec}
This paper has provided what is, to the best of our knowledge, a first unified study on the use of seven different ML-based object detection and tracking algorithms in the context of underwater ROV control. First, after briefly reviewing the related works, this paper has provided a detailed description of seven popular ML-based object detection and tracking algorithms that have often been used in conventional computer vision applications, but rarely used in the underwater context. Then, this paper has described the design of our automated ROV position locking system built around the one-shot object trackers feeding their output bounding box deviations into a PI controller. Extensive real-world experiments were conducted using the proposed position locking pipeline which controls in real-time a BlueROV2 platform within a swimming pool environment. Our experiments clearly show that the \textit{Channel and Spatial Reliability Tracker (CSRT)} outperforms other methods for the design of underwater ROV control systems. 

\section*{Acknowledgement}

\footnotesize{
\bibliographystyle{IEEEtran}
\bibliography{References}
}

\begin{IEEEbiography}[{\includegraphics[width=1in,height=1.5in,clip,keepaspectratio]{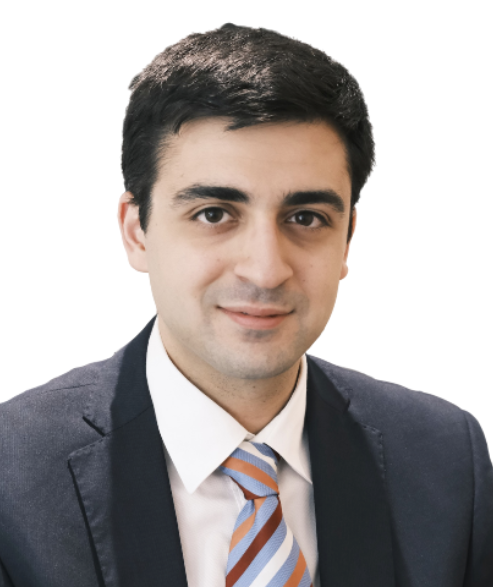}}]{Ali Safa}
(Member, IEEE) is currently an Assistant Professor with the College of Science and Engineering, Hamad Bin Khalifa University (HBKU), Doha, Qatar. He received the M.Sc. degree in electrical engineering from the Université Libre de Bruxelles, Brussels, Belgium, and the Ph.D. degree in AI-driven processing for extreme edge applications from the KU Leuven, Belgium, in collaboration with the IMEC R\&D Center, Leuven Belgium. He has been a Visiting Researcher with the University of California at San Diego (UC San Diego), La Jolla, USA, in Spring 2023. He has authored and co-authored more than 40 research articles in international journals and conferences, and is the author of a Springer book on the application of Neuromorphic Computing to Sensor Fusion and Drone Navigation tasks. His research interests include Edge AI, continual learning, and sensor fusion for robot perception. 
\end{IEEEbiography}
\begin{IEEEbiography}
[{\includegraphics[width=1in,height=1.5in,clip,keepaspectratio]{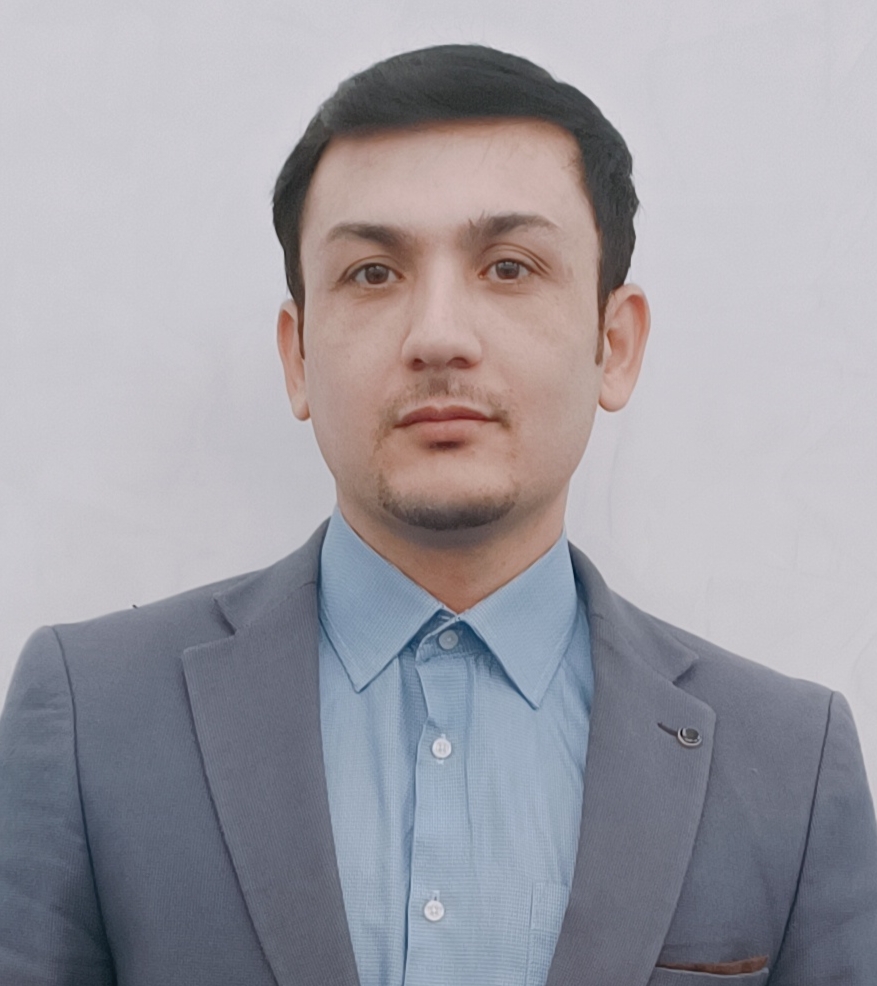}}]{Waqas Aman}
(Member, IEEE) is currently a Postdoctoral Fellow at the College of Science and Engineering, Hamad Bin Khalifa University, Qatar. He earned his Ph.D. in Electrical Engineering from the Information Technology University, Lahore, Pakistan, with a focus on security and reliability in underwater acoustic communication networks. During his Ph.D. studies, he was a visiting researcher at the University of Glasgow, UK, and a research intern at KAUST, Saudi Arabia for six months each.  His work has been recognized with prestigious awards, including the Best Paper Award at IEEE ISNCC 2023, and the top downloaded technical paper in 2018-2019 by Trans. on ETT, Wiley. So far, he has published 23 research articles in international journals and conferences. 
\end{IEEEbiography}
\begin{IEEEbiography}[{\includegraphics[width=1in,height=1.5in,clip,keepaspectratio]{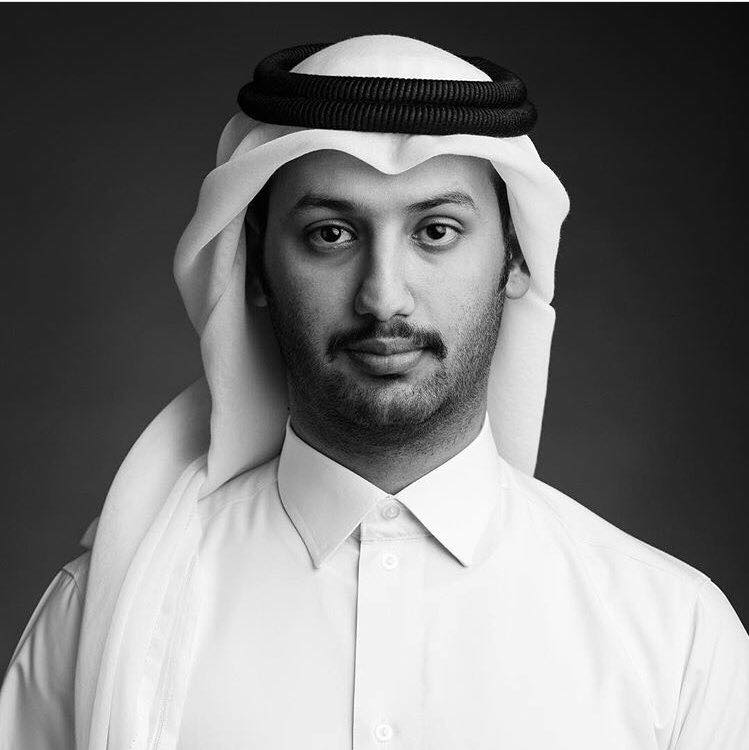}}]{Ali Al-Zawqari}
(Member, IEEE) received the M.Sc. degree in electrical engineering from the Bruface program, Brussels, Belgium, and the Ph.D. degree in enhancing generalization and fairness in machine learning from the Vrije Universiteit Brussel, Belgium. He is a senior fellow with the QatarDebate Center, Doha, Qatar. He is also a postdoctoral research associate at the ELEC department at Vrije Universiteit Brussel, Belgium. He has authored and co-authored more than 15 research articles in international journals and conferences. His research interests include digital signal processing, fundamental machine learning, graph theory, human Learning, and Arabic NLP. 
\end{IEEEbiography}

\begin{IEEEbiography}
[{\includegraphics[width=1in,height=1.5in,clip,keepaspectratio]{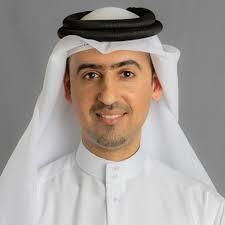}}]{Saif Al-Kuwari}
(Senior Member, IEEE)  is an Associate Professor in the Faculty of
Science and Engineering of HBKU. He received a BEng in
Computers and Networks from the University of Essex (UK)
in 2006 and two PhDs from the University of Bath and Royal
Holloway, University of London (UK) in Computer Science
in 2012. His research interests include applied cryptography,
Quantum Computing, Computational Forensics, and their
connections with Machine Learning. He is an IET and BCS
fellow and an IEEE and ACM senior member
\end{IEEEbiography}








\end{document}